\documentclass[runningheads]{llncs}
\usepackage{graphicx}
\usepackage{amssymb, amsmath, bm, latexsym,comment}
\usepackage{multirow}
\usepackage{xcolor}
\usepackage{subfigure}
\usepackage{indentfirst}
\usepackage{setspace}
\usepackage{verbatim}
\usepackage{array}
\usepackage{booktabs}
\usepackage{bbding}
\usepackage[misc]{ifsym} 

\providecommand{\Zxhreftb}[1]{Table~\ref{#1}}
\providecommand{\zxhreftb}[1]{Table~\ref{#1}}
\providecommand{\zxhreffig}[1]{Fig.~\ref{#1}}

\providecommand{\citep}[1]{\cite{#1}}

\begin{document}
\title{Random Style Transfer based Domain Generalization Networks Integrating Shape and Spatial Information} 
\titlerunning{STDGNs with shape and spatial information}

\author{Lei Li\inst{1, 2, 3} \and
Veronika A. Zimmer \inst{3} \and 
Wangbin Ding  \inst{4} \and 
Fuping Wu \inst{2, 5} \and 
Liqin Huang \inst{4} \and 
Julia A. Schnabel \inst{3} \and 
Xiahai Zhuang\inst{2} ${^{(\textrm{\Letter})}}$ \\
}
\authorrunning{L. Li et al.}

\institute{School of Biomedical Engineering, Shanghai Jiao Tong University, Shanghai, China \and
School of Data Science, Fudan University, Shanghai, China \\
\email{zxh@fudan.edu.cn} \and 
School of Biomedical Engineering and Imaging Sciences, King’s College London, London, UK\and
College of Physics and Information Engineering, Fuzhou University, Fuzhou, China\and
School of Management, Fudan University, Shanghai, China
}

\maketitle 
\begin{abstract}
Deep learning (DL)-based models have demonstrated good performance in medical image segmentation.
However, the models trained on a known dataset often fail when performed on an unseen dataset collected from different centers, vendors and disease populations.
In this work, we present a random style transfer network to tackle the domain generalization problem for multi-vendor and center cardiac image segmentation.
Style transfer is used to generate training data with a wider distribution/ heterogeneity, namely domain augmentation.
As the target domain could be unknown, we randomly generate a modality vector for the target modality in the style transfer stage, to simulate the domain shift for unknown domains.
The model can be trained in a semi-supervised manner by simultaneously optimizing a supervised segmentation and a unsupervised style translation objective.
Besides, the framework incorporates the spatial information and shape prior of the target by introducing two regularization terms.
We evaluated the proposed framework on 40 subjects from the M\&Ms challenge2020, and obtained promising performance in the segmentation for data from unknown vendors and centers.

\end{abstract}

\section{Introduction}
Quantification of volumetric changes during the cardiac cycle is essential in the diagnosis and therapy of cardiac diseases, such as dilated and hypertrophic cardiomyopathy.
Cine magnetic resonance image (MRI) can capture cardiac motions and presents clear anatomical boundaries.
For quantification, accurate segmentation of the ventricular cavities and myocardium from cine MRI is generally required \citep{journal/PAMI/zhuang2018}.

Recently, deep learning (DL) based methods have obtained promising results in cardiac image segmentation.
However, the generalization capability of the DL-based models is limited, i.e., the performance of a trained model will be drastically degraded for unseen data.
It is mainly due to the existence of the domain shift or distribution shift, which is common among the data collected from different centers and vendors.
To tackle this problem, domain adaptation and domain generalization are two feasible schemes in the literature.
Domain adaptation algorithms normally assume that the target domain is accessible in the training stage.
The models will be trained using labeled source data and a few labeled or unlabeled target data, and learn to align the source and target data onto a domain-invariant feature space. 
For example, Yan et al. adopted an unpaired generative adversarial network (GAN) for vendor adaptation without using manual annotation of target data \citep{conf/MICCAI/yan2019}.
Zhu et al. proposed a boundary-weight domain adaptive network for prostate MRI segmentation \citep{journal/TMI/zhu2019}.
Domain generalization, on the other hand, is more challenging due to the absence of target information.
It aims to learn a model that can be trained using multi-domain source data and then directly generalized to an unseen domain without retraining.
In clinical scenarios, it is impractical to retrain a model each time for new domain data collected from new vendors or centers.
Therefore, improving the generalization ability of trained models would be of great practical value.

Existing domain generalization methods can be summarized into three categories, 
1) domain-invariant feature learning approaches, which focus on extracting task-specific but domain-invariant features \citep{conf/CVPR/li2018,conf/ECCV/li2018}; 
2) model-agnostic meta-learning algorithms, which optimize on the meta-train and meta-test domain split from the available source domain \citep{conf/NIPS/dou2019,conf/arXiv/liu2020}; 
3) data augmentation methods, which enlarge the scope of distribution learned by models for model generalization \citep{journal/Arxiv/chen2019,journal/TMI/zhang2020}. 
For example, Volpi et al. utilized adversarial data augmentation to iteratively augment the dataset with examples from a fictitious target domain \citep{conf/ANIPS/volpi2018}.
Zhang et al. propose a deep stacked data augmentation (BigAug) approach to simulate the domain shift for domain generalization \citep{journal/TMI/zhang2020}.
Chen et al. employed some data normalization and augmentation for cross-scanner and cross-site cardiac MR image segmentation \citep{journal/Arxiv/chen2019}.
However, data augmentation in the source domain is not enough to mitigate the generalization gap, as the domain shift is not accidental but systematic \citep{conf/MICCAI/yan2019}.

In this work, we propose novel random style transfer based domain generalization networks (STDGNs) incorporating spatial and shape information.
The framework focuses on improving the networks’ ability to deal with the modality-level difference instead of structure-level variations, i.e., domain augmentation.
Specifically, we first transfer the source domain images into random styles via the style transfer network \citep{journal/MedIA/yuan2020}.
The generated new domain and known source domain images are then employed to train the segmentation network.
We combine the two steps by connecting a style transfer and style recover network, to constrain the two domains with the same structure but different styles.
With the domain augmentation strategy, the segmentation network is supposed to be domain-invariant and therefore can be adapted to unknown target domains.
The framework can be trained in a semi-supervised manner, as it consists of an unsupervised style transfer task and a supervised segmentation task.
Besides, we employ a shape reconstruction network (SRN) to include shape priors as a constrain, and utilize a spatial constrain network (SCN) to consider the large variety of 2D slices from various slice positions of 3D cardiac images \citep{conf/MICCAI/yue2019}.

\begin{figure*}[t]\center
 \includegraphics[width=0.98\textwidth]{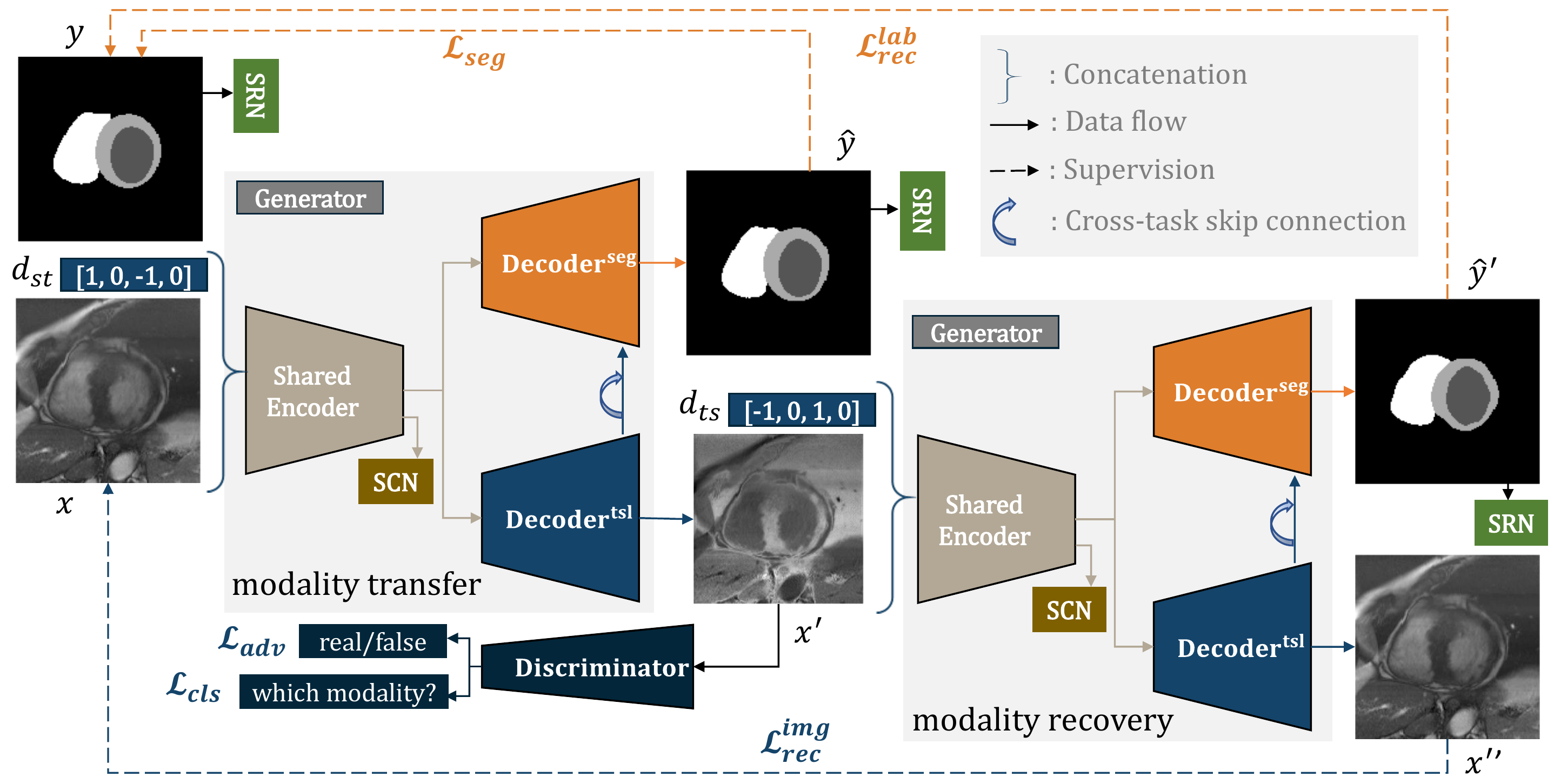}\\[-2ex]
   \caption{The proposed STDGNs incorporating shape and spatial information. The generator is consist of one shared encoder and two decoders for segmentation and style translation. Note that the skip connections between the encoder and two decoders are omitted here.}
\label{fig:method:framework}\end{figure*}

\section{Method} \label{method}
\zxhreffig{fig:method:framework} provides an overview of the proposed framework.
The proposed network is a cascaded modified U-Net consisting of two decoders for cardiac segmentation and style transfer, respectively.
The generators consist of a shared encoder and two specific decoders, while the discriminator includes two parts, i.e., real/ fake discriminator and modality discriminator.
Section \ref{method:DG} presents the unified segmentation and style transfer framework based on GANs.
To use unlabeled data, semi-supervised learning is employed, followed by a cross-task attention module to efficiently connect the segmentation and style transfer tasks (see Section \ref{method:SSL}).
In Section \ref{method:SE}, two losses based on the shape and spatial information are introduced as regularization terms.

\subsection{Deep Domain Generalization Framework}\label{method:DG}
Given $N$ source domains $\mathcal{D}_s=\left\{\mathcal{D}_{1}, \mathcal{D}_{2}, \cdots, \mathcal{D}_{N}\right\}$, we aim to construct a model $f(X)$,  
\begin{equation}
f(X) \rightarrow Y, X \in \mathcal{D}_s \cup \mathcal{D}_{t},
\end{equation}
where $\mathcal{D}_{t}$ are unknown target domains, and $X,Y \in \mathbb{R}^{1 \times H \times W}$ denote the image set and corresponding segmentation set.
To achieve this, we propose to employ the style transfer to randomly generate new domains $\mathcal{D}_{s}^{\prime}$ with the same structure as the input source domain.
Specifically, we assign a one-hot encoding modality vector $v \in[0,1]^{N+1}$ for source domain image $x \in \mathbb{R}^{1 \times H \times W}$, and then randomly generate a modality vector $v^{\prime } \in[0,1]^{N+1}$ for target domain image.
The modality difference vector $d_{st}=v^{\prime}-v, d_{st} \in[-1,0,1]^{k}$ is broadcasted from $\mathbb{R}^{k}$ to $\mathbb{R}^{k \times H \times W}$ and then concatenated with the input image. 

As \zxhreffig{fig:method:framework} shows, in the modality transfer phase, the generator $G$ translates the source domain image $x$ into a new target domain image $x^{\prime}$ and predicts the segmentation $\hat{y}$, i.e., $G\left(x, d_{s t}\right) \rightarrow\left(x^{\prime}, \hat{y}\right)$.
The generated $x^{\prime}$ is then inputted into a discriminator $D={D_\text{src}, D_\text{cls}}$, where $D_\text{src}$ recognizes it as real/ fake and $D_\text{cls}$ predicts its modality.
In the modality recovery phase, $d_{ts}=v-v^{\prime}$ is concatenated with $x^{\prime}$ as the input of $G$, to obtain the reconstructed original source image $x^{\prime\prime}$ and its label $\hat{y}^{\prime}$ for structure preservation. 
Therefore, the adversarial loss of the GAN is adopted as follows,
\begin{equation}
\mathcal{L}_{adv}^{D}=\mathbb{E}_{x}\left[D_{src}(x)\right]-\mathbb{E}_{x^{\prime}}\left[D_{src}\left(x^{\prime}\right)\right]-\lambda_{gp} \mathbb{E}_{\hat{x}}\left[\left(\left\|\nabla_{\hat{x}} D_{src}(\hat{x})\right\|_{2}-1\right)^{2}\right],
\end{equation}
\begin{equation}
\mathcal{L}_{adv}^{G}=\mathbb{E}_{x^{\prime}}\left[D_{src}\left(x^{\prime}\right)\right],
\end{equation}
where $\hat{x}$ is sampled from the real and fake images and $\lambda_{gp}$ is the balancing parameter of gradient penalty.
For style transfer, $D_{cls}$ learns to classify $x$ into its source modality $v$ based on $\mathcal{L}_{cls}^{r}$, while $G$ tries to transfer source modality into target modality $v^{\prime}$ via $\mathcal{L}_{cls}^{f}$ with
\begin{equation}
\mathcal{L}_{cls}^{r}=\mathbb{E}_{x, v}\left[-\log D_{cls}(v \mid x)\right],
\end{equation}
\begin{equation}
\mathcal{L}_{cls}^{f}=\mathbb{E}_{x^{\prime}, v^{\prime}}\left[-\log D_{cls}\left(v^{\prime} \mid x^{\prime}\right)\right].
\end{equation}
The cycle consistency losses include $\mathcal{L}_{rec}^{lab}$ and $\mathcal{L}_{rec}^{img}$. $\mathcal{L}_{rec}^{img}$ is defined as
\begin{equation}
\mathcal{L}_{rec}^{img}=\mathbb{E}_{x, x^{\prime \prime}}\left[\left\|x-x^{\prime \prime}\right\|_{1}\right],
\end{equation}
where $x^{\prime \prime}$ is the predicted source domain image. $\mathcal{L}_{rec}^{lab}$ is formulated in Section \ref{method:SE}. 
Hence, the final overall loss to optimize the generator and discriminator are defined as,
\begin{equation}
\min _{D} \mathcal{L}_{D}=-\mathcal{L}_{adv}^{D}+\lambda_{cls} \mathcal{L}_{cls}^{r},
\end{equation}
\begin{equation}
\min _{G} \mathcal{L}_{G}=-\mathcal{L}_{adv}^{G}+\lambda_{cls} \mathcal{L}_{cls}^{f}+\lambda_{rec}^{img} \mathcal{L}_{rec}^{img}+\lambda_{seg} \mathcal{L}_{seg}+\lambda_{rec}^{lab} \mathcal{L}_{rec}^{lab},
\end{equation}
where all $\lambda$ are balancing parameters, and $\mathcal{L}_{seg}$ and $\mathcal{L}_{rec}^{lab}$ are introduced in Section \ref{method:SE}.

\subsection{Semi-supervised Learning with a Cross-task Attention}  \label{method:SSL}
To employ unlabeled data, we utilize a semi-supervised method via multi-task attention.
Specifically, the proposed two-task networks include a supervised segmentation task and an unsupervised style transfer task, and the two tasks are connected via an attention mechanism.
The style transfer decoder is trained without requirements to manually labeled data, and therefore the model can learn more discriminating features for segmentation from unlabeled images. 

\begin{figure*}[!t]\center
	\subfigure[] {\includegraphics[width=0.48\textwidth]{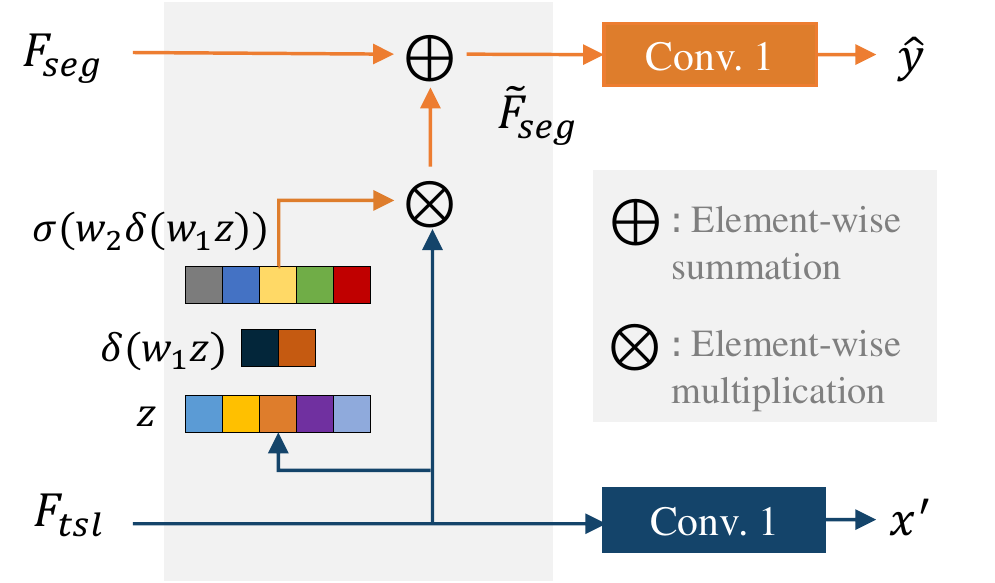}}
	\subfigure[] {\includegraphics[width=0.48\textwidth]{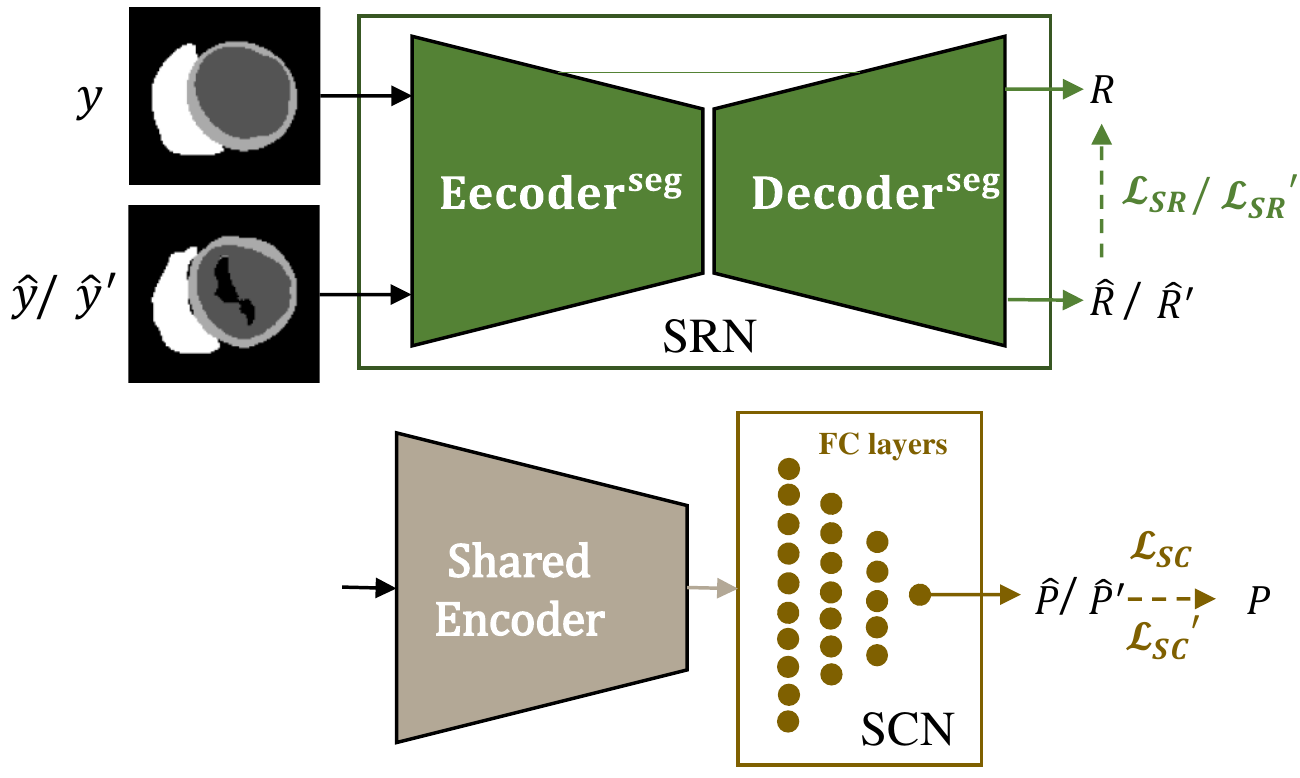}}
	\caption{
		Two auxiliary modules for the proposed framework: 
		(a) cross-task skip connection module: the auxiliary features from the style transfer task $F_{tsl}$ should be recalibrated before connected with the features of the segmentation task $F_{seg}$.
		(b) shape reconstruction and spatial constrain module: SRN reconstructs the label images to encode the shape prior information, and SCN predict the slice positions to incorporate the spatial information. }
	\label{fig:method:module}\end{figure*}

The assumption is that combining the features from the style transfer task can be beneficial to encode more representative and robust structural information for the segmentation task.
However, due to the diverse feature spaces of the two tasks, a feature recalibration module is required to discard unrelated information.
As \zxhreffig{fig:method:module} (a) shows, we employ a cross-task skip connection for the last layers of the two decoders based on the squeeze and excitation (SE) module \citep{conf/CVPR/hu2018}. 
The recalibration feature is defined as follows,
\begin{equation}
\tilde{F}_{seg}=F_{seg}+F_{tsl} \otimes \sigma\left(W_{2} \delta\left(W_{1} z\right)\right),
\end{equation}
where $F_{seg}$ and $F_{tsl}$ are the feature maps before the final output layers of the two tasks, $W_{1}$ and $W_{2}$ are fully connected layers, and $z=$ AvgPool $\left(F_{tsl}\right) \in \mathbb{R}^{C \times 1 \times 1}$.
Specifically, $W_{1}$ is employed to squeeze the spatial information into a channel vector, i.e., $W_{1} z \in \mathbb{R}^{r \times 1 \times 1}$, while $W_{2}$ is used to recover the channel number from $r$ to $C$ for attention.

\subsection{Shape Reconstruction and Spatial Constraint for Segmentation} \label{method:SE}
To employ the shape prior, we use the SRN to reconstruct the label image to learn an intermediate representation, see \zxhreffig{fig:method:module} (b).
Specifically, the SRN model is pre-trained using available label images and corresponding predicted label images from basic U-Net \citep{conf/MICCAI/ronneberger2015}.
It can be regarded as a constraint to regularize the predicted segmentation into a desired realistic shape. 
As \zxhreffig{fig:method:framework} shows, SRN is connected with the output of Decoder$^\text{seg}$ to refine the shape of the prediction.
The loss function of SRN is defined as,
\begin{equation}
\mathcal L_{SR}=\mathbb{E}_{y, \hat{y}, \hat{y}^{\prime }}\left[\left\|R-(\hat{R} \text{ or } \hat{R}^{\prime }) \right\|_{2}\right],
\end{equation}
where $R$ denotes the reconstructed gold standard label from $y$, while $\hat{R}$ and $\hat{R}^{\prime}$ indicate the reconstructed predicted label from $\hat{y}$ and $\hat{y}^{\prime}$, respectively.

Considering the appearance of basal and apical slices can differ significantly, we utilize the SCN to predict the spatial information of each slice, i.e, its normalized distance to the central slice.
As \zxhreffig{fig:method:module} (b) shows, the SCN is connected to the bottom of U-Net via fully connected (FC) layers to predict the slice position.
The SC loss is formulated as,
\begin{equation}
\mathcal L_{SC}=\mathbb{E}_{x, x^{\prime }}\left[\left\|P-(\hat{P}\text{ or } \hat{P}^{\prime }) \right\|_{2}\right],
\end{equation}
where $P$ is the ground truth position, while $\hat{P}$ and $\hat{P}^{\prime}$ are the predicted positions of $x$ and $x^{\prime}$, respectively.
By incorporating the SR and SC loss, the segmentation loss and the label reconstruction loss both can be defined as,
\begin{equation}
\mathcal L_{seg}=L_{rec}^{lab}=\mathcal L_{CR} + \lambda_{SR} \mathcal L_{SR} + \lambda_{SC} \mathcal L_{SC},
\end{equation}
where $\mathcal L_{CR}$ is the cross-entropy loss, and $\lambda_{SR}$ and $\lambda_{SC}$ are the balancing parameters.

\begin{table} [t] \center
    \caption{
        The distribution of training and test data. $I^\mathrm{lab}$: images with gold standard labels; $I^\mathrm{unl}$: images without labels.
     }
\label{tb:exp:dataset}
{\small
\begin{tabular}{ l| l l | l l *{5}{@{\ \,} l }}\hline
\multirow{2}*{Vendor}& \multicolumn{2}{c}{Training data} & \multicolumn{2}{c}{Test data}\\
\cline{2-5}
~  & center & \quad cine MRI   & \quad center & \quad cine MRI \\
\hline
\quad A&  1&  \quad 75 $I^\mathrm{lab}$          &\quad 1/ 6&  \quad 5/ 5 $I^\mathrm{unl}$\\  
\quad B&  2/ 3&  \quad 50/ 25 $I^\mathrm{lab}$   &\quad 2/ 3&  \quad 5/ 5 $I^\mathrm{unl}$\\ 
\quad C&  4&  \quad 25 $I^\mathrm{unl}$          &\quad 4&  \quad 10 $I^\mathrm{unl}$\\ 
\quad D&  N/A &  \quad N/A                       &\quad 5&  \quad 10 $I^\mathrm{unl}$\\ 
\hline
\end{tabular} }\\
\end{table}

\section{Experiments}

\subsection{Materials}
\subsubsection{Data Acquisition and Pre-processing.}
The data is from the M\&Ms2020 challenge~\cite{link/MMs2020}, which provides 375 short-axis cardiac MRIs from patients with hypertrophic and dilated cardiomyopathies as well as healthy subjects. 
All subjects were collected from three different countries (Spain, Germany and Canada) using four different MRI vendors (Siemens, General Electric, Philips and Canon).
As \zxhreftb{tb:exp:dataset} shows, the training data includes 150 labeled images from vendor A and B and 25 unlabeled images from vendor C. 
The test data in this study contains 40 images from each vendor, i.e., A, B, C and a novel unseen vendor D. 
All slices were cropped into a unified size of $144\!\times\!144$ centering at the heart region and were normalized using z-score.

\subsubsection{Gold Standard and Evaluation.}
The images have been manually segmented by experienced clinicians with labels of the left ventricle (LV), right ventricle (RV) blood pools, and left ventricular myocardium (Myo).
Note that manual annotations are provided for only two cardiac time frames, i.e., end-diastole (ED) and end-systole (ES).
These manual segmentations were considered as the gold standard.
For segmentation evaluation, Dice volume overlap and Hausdorff distance (HD) were applied.

\subsubsection{Implementation.}
The framework was implemented in PyTorch, running on a computer with 1.90 GHz Intel(R) Xeon(R) E5-2620 CPU and an NVIDIA TITAN X GPU. 
We used the SGD optimizer to update the network parameters (weight decay=0.0001, momentum=0.9). 
The initial learning rate was set to 3e-4 and divided by 10 every 5000 iterations, and batch size was set to 20. 
Note that the generator and discriminator are alternatively optimized every 50 epochs.
The balancing parameters were set as follows, $\lambda_{seg}=100$, $\lambda_{rec}^{img}=100$, $\lambda_{SC}=1$, $\lambda_{SR}=100$, $\lambda_{gp}=10$, $\lambda_{cls}=10$, and  $\lambda_{rec}^{lab}$ was initial set to 0 and then gradually increase to 100.

\begin{table*} [t] \center
    \caption{
    Summary of the quantitative evaluation results of cardiac image segmentation on the test data including four vendors. 
    SR and SC refer to the two auxiliary regularization terms; semi denotes that the model employed both labeled and unlabeled data.
     }
\label{tb:result:Dice}
{\small
\begin{tabular}{ l| l l l | l  *{5}{@{\ \,} l }}\hline
Method    & \quad Dice (LV)  & \quad Dice (Myo) & \quad Dice (RV) & \quad Mean\\
\hline
U-Net &$ 0.700 \pm 0.322 $&  $ 0.675 \pm 0.272 $&  $ 0.590 \pm 0.369 $ &  $ 0.655 \pm 0.326 $\\ 
MASSL$^\text{semi}$&$0.727 \pm 0.292 $& $ 0.691 \pm 0.253 $&  $ 0.567 \pm 0.371 $ &  $ 0.662 \pm 0.315 $\\ 
\hline\hline
STDGNs$^\text{semi}$ &$ 0.738 \pm 0.277 $& $ 0.694 \pm 0.238 $& $ 0.601 \pm 0.356 $ & $ 0.678 \pm 0.299 $\\
STDGNs$^\text{semi}$-SR   &$ 0.754 \pm 0.302 $&  \bm{$ 0.718 \pm 0.261 $}&  $ 0.619 \pm 0.370 $ &  $ 0.697 \pm 0.318 $\\  
STDGNs$^\text{semi}$-SRSC &\bm{$ 0.767 \pm 0.256 $}&  $ 0.716 \pm 0.237 $& \bm{$ 0.636 \pm 0.355 $} & \bm{$ 0.706 \pm 0.292 $}\\
\hline
\end{tabular} }\\
\end{table*}

\subsection{Result}

\subsubsection{Comparisons with Literature and Ablation Study.}
\Zxhreftb{tb:result:Dice} presents the quantitative results of different methods for cardiac segmentation in terms of Dice.
Two algorithms were used as baselines for comparisons, i.e., fully supervised U-Net \citep{conf/MICCAI/ronneberger2015} and multi-task attention-based semi-supervised learning (MASSL) network \citep{conf/MICCAI/chen2019}.
MASSL is a two-task network consisting of two decoders, which are used for segmentation prediction and image reconstruction, respectively.
Compared to U-Net, MASSL employed the unlabeled data for training, and therefore obtained better mean Dice though with worse Dice on RV.
STDGNs is also semi-supervised and it fully utilized the supervised and unsupervised task relationship via a cross-task attention module.
It obtained evidently better Dice scores in all sub-regions compared to U-Net and MASSL.

To explore the effectiveness of the proposed shape and spatial constrain, we also adopted an ablation study.
One can see that STDGNs-SR outperformed STDGNs in terms of Dice, and incorporating the SC module further improved the performance on LV and RV segmentation. 
Furthermore, one can see that the performance on the RV segmentation were all worse than that on the LV and Myo for these algorithms.
It may due to the fact that the shape of the RV is generally more irregular than that of the LV and Myo.

\begin{figure*}[!t]\center
	\subfigure[] {\includegraphics[width=0.49\textwidth]{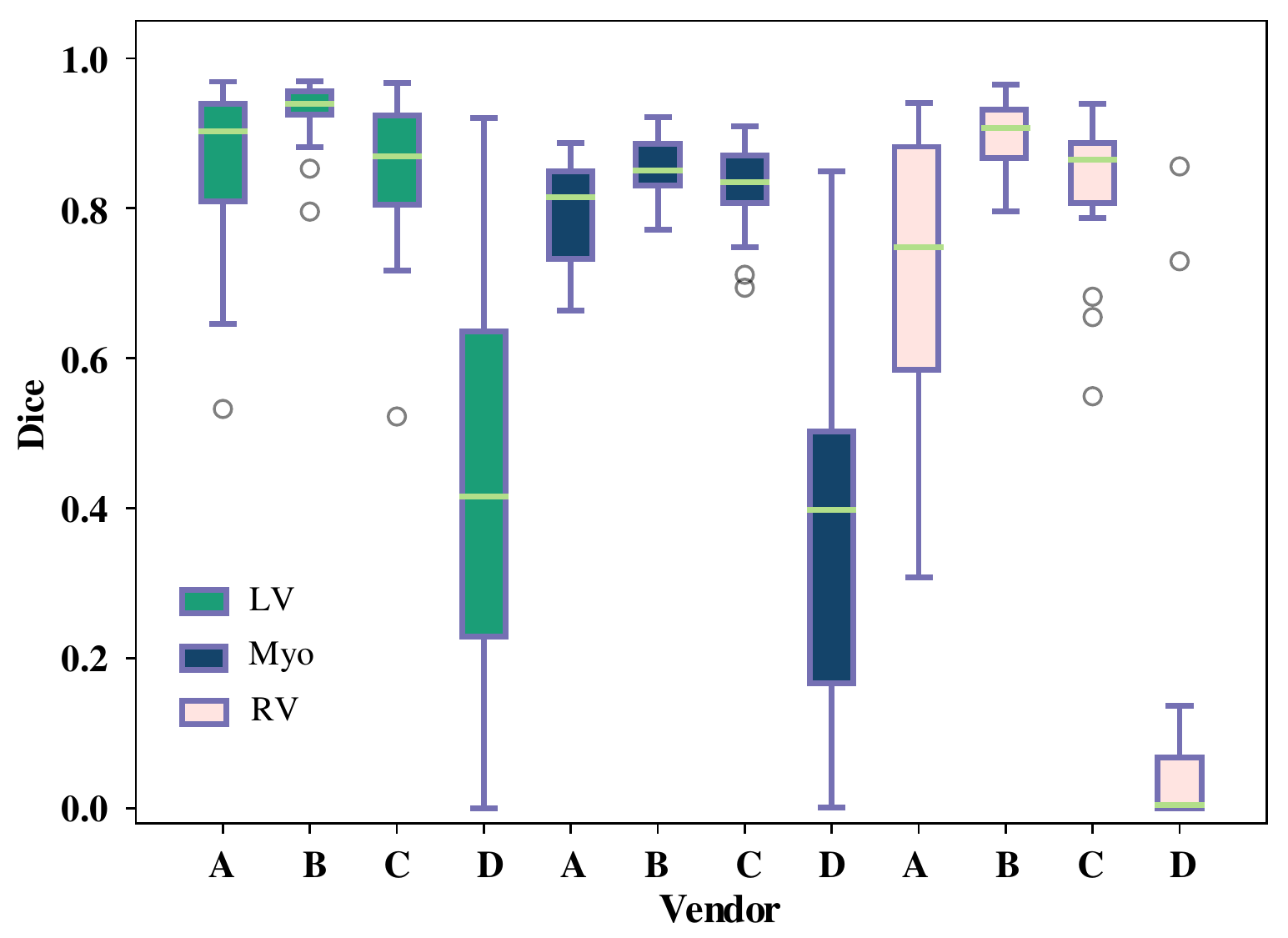}}
	\subfigure[] {\includegraphics[width=0.49\textwidth]{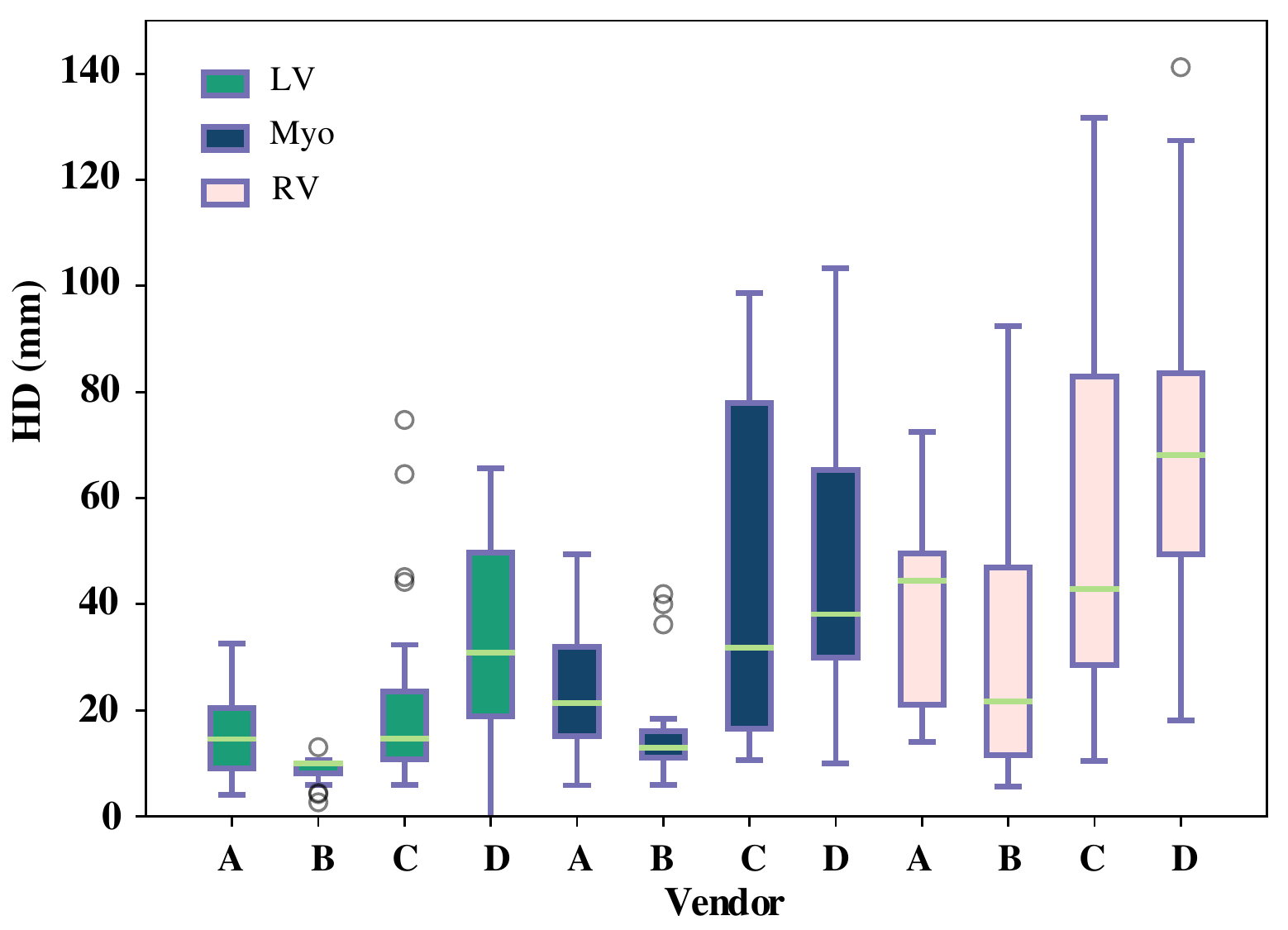}}
	\subfigure[] {\includegraphics[width=0.49\textwidth]{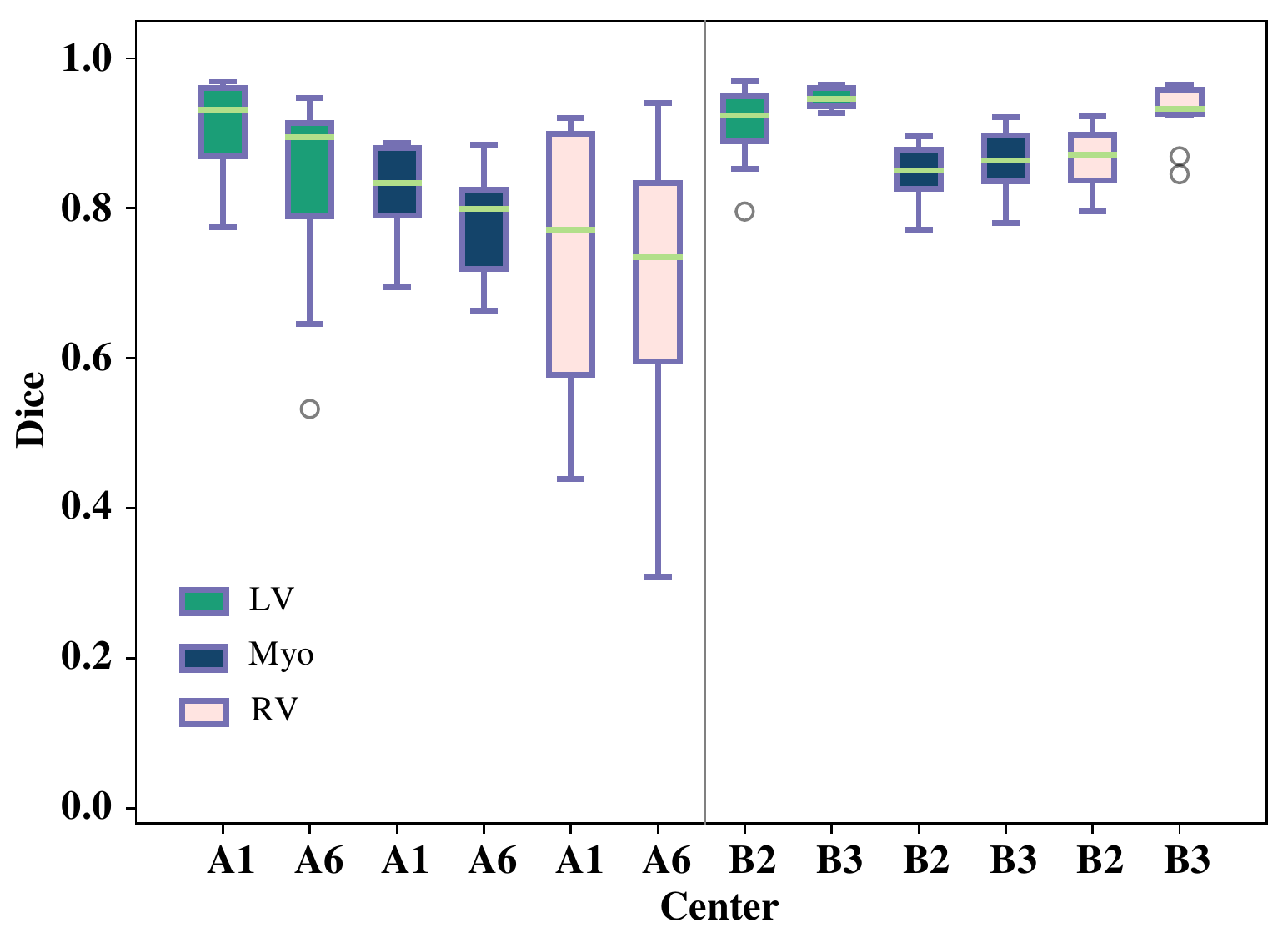}}
	\subfigure[] {\includegraphics[width=0.49\textwidth]{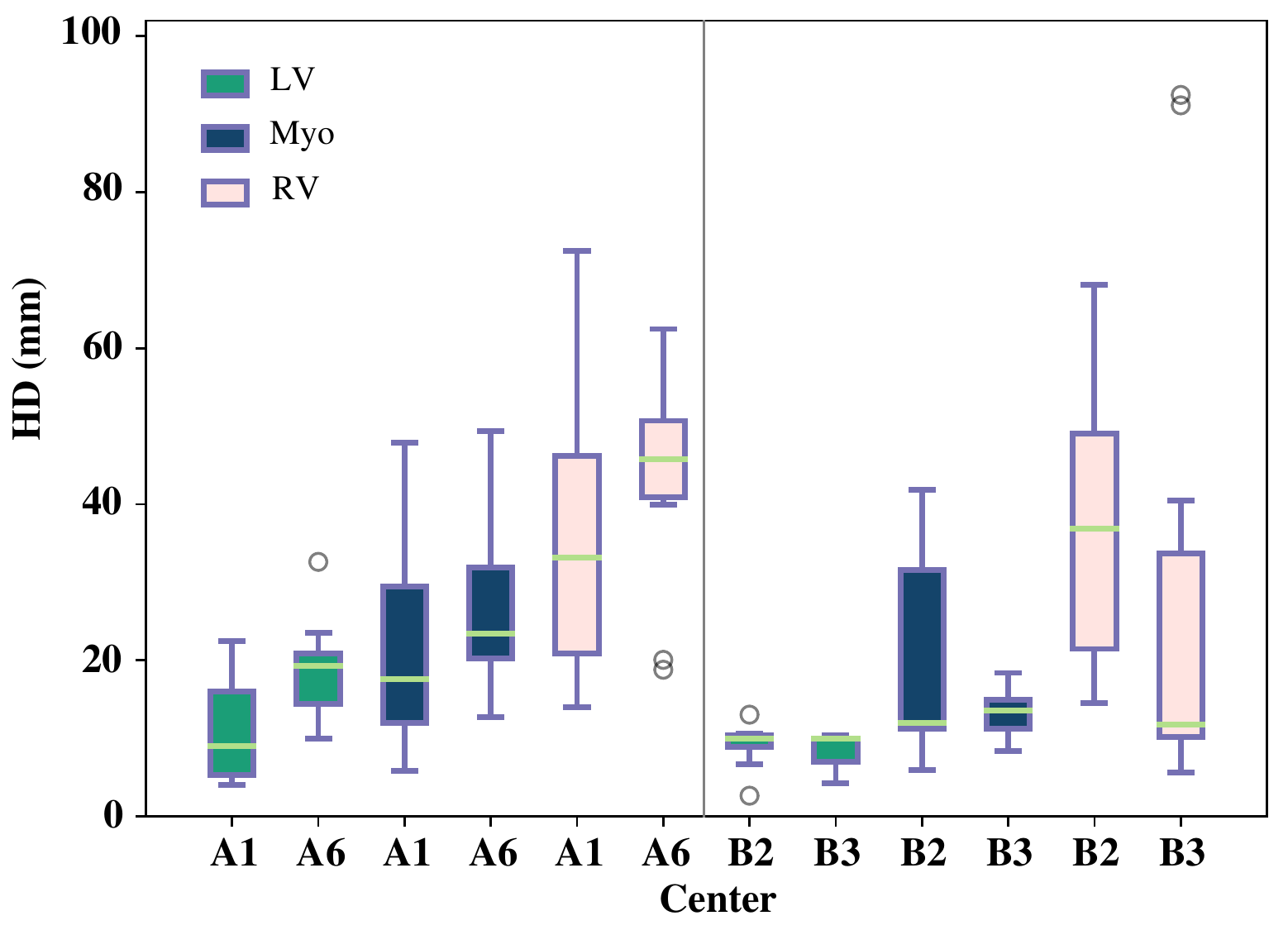}}
	\caption{
		The quantification results of the three substructures by the proposed method on each vendor and center: 
		(a) Dice of the four vendors;
		(b) HD of the four vendors;
		(c) Dice of the two centers on vendor A and B, respectively;
		(d) HD of the two centers on vendor A and B, respectively. 
		Here, A1 and A6 refer to data of center 1 and 6 from vendor A, while B2 and B3 denote the data of center 2 and 3 from vendor B.
		}
	\label{fig:result:boxplot}
\end{figure*}

\subsubsection{Performance on the Data from Different Vendors}
To investigate the generalization ability of our model, we compared the performances on data from various vendors and centers.
\zxhreffig{fig:result:boxplot} (a) and (b) illustrate the Dice and HD of the three substructures on each vendor.
It is evident that the model performed generally well on the known domains, i.e., vendor A, B and C.
Though the label images of vendor C are unavailable, our model still obtained promising results, thanks to the semi-supervised learning.
However, for the unknown domain vendor D, the performance drastically dropped, especially for the RV segmentation.
To verify the generalization performance of the proposed model, we further compared the results of our model with that of U-Net on vendor D.
As \Zxhreftb{tb:result:Vendor_D} shows, the proposed model performed evidently better in generalization compared to the baseline, though there is still a huge scope for improvements.

\subsubsection{Performance on the Data from Different Centers.}
Besides the generalization on different vendors, we also explored the effectiveness of the proposed model on different centers.
As vendors C and D include one single center, we only present the results of two centers of vendor A and B, respectively (see \zxhreffig{fig:result:boxplot} (c) and (d)).
One can see that the results of center A6 were generally worse than that of center A1, as the data of center A6 is not included in the training dataset.
However, compared to the performance decline on the new vendor D, the performance decrease on the new center is evidently less.
For the centers already included in the training dataset (centers B2 and B3), the model obtained promising results in terms of both Dice and HD.
Note that the performance on center B3 is evidently better than that of center B2, though the training data includes more center B2 data (see \Zxhreftb{tb:exp:dataset}).
This may be because the center B2 data is more challenging to segment or center B3 has more similar appearance with other centers in the training data.

\begin{table*} [t] \center
    \caption{
    The generalization performance of the baseline and proposed algorithm on the unknown domain vendor D. 
     }
\label{tb:result:Vendor_D}
{\small
\begin{tabular}{ l| l l l  *{5}{@{\ \,} l }}\hline
Method    & \quad Dice (LV)  & \quad Dice (Myo) & \quad Dice (RV)\\
\hline
U-Net &\quad $ 0.241 \pm 0.284 $&  \quad $ 0.277 \pm 0.253 $& \quad $ 0.063 \pm 0.183 $ \\ 
STDGNs$^\text{semi}$-SRSC & \quad $ 0.425 \pm 0.282 $ &  \quad $ 0.380 \pm 0.103 $& \quad $ 0.240 \pm 0.355 $ \\
\hline
\end{tabular} }\\
\end{table*}

\subsubsection{Performance on the ED and ES Phase.}
\Zxhreftb{tb:result:ED_ES} presents the quantitative results of the proposed method for cardiac segmentation at the ED and ES phase.
One can see that the performance was substantially higher in ED than in ES for all substructures in terms of Dice, but was similar in HD.
This might be caused by the larger ventricular shape in the ED phase than the ES phase.
 
\begin{table*} [t] \center
    \caption{
    Dice and HD for cardiac image segmentation at ED and ES phase.
     }
\label{tb:result:ED_ES}
{\small
\begin{tabular}{ l| l l l |  l l l *{7}{@{\ \,} l }}\hline
\multirow{2}*{Phase}& \multicolumn{3}{c}{Dice} & \multicolumn{3}{c}{HD (mm)}\\
\cline{2-7}
~  & \quad LV & \quad Myo & \quad RV & \quad LV & \quad  Myo & \quad RV \\
\hline
ED &  $ 0.818 \pm 0.236 $&   $ 0.724 \pm 0.209 $&  $ 0.671 \pm 0.361 $ &  $ 19.9 \pm 17.7 $ &  $ 32.8 \pm 23.1 $ &  $ 51.0 \pm 32.5 $ \\ 
ES &  $ 0.715 \pm 0.268 $&   $ 0.708 \pm 0.265 $&  $ 0.601 \pm 0.351 $ &  $ 20.3 \pm 16.4 $ &  $ 33.3 \pm 27.2 $ &  $ 50.4 \pm 34.6 $ \\ 
\hline
\end{tabular} }\\
\end{table*}

\section{Conclusion}
In this work, we have proposed an end-to-end learning framework for domain generalization via random style transfer.
The proposed algorithm has been applied to 40 cine images and obtained promising results on the unknown domain images from new vendors and centers.
Especially, the model could generalize better on new centers than new vendors.
The results have demonstrated the effectiveness of the proposed spatial and shape constrain schemes, which encourage the network to learn more domain-invariant features.
Our technique can be easily extended to other segmentation tasks on data from new vendors and centers.
A limitation of this work is that the definition of the modality vector requires the number of domains, which is normally unknown.
Besides, the style transfer only can modify modality globally, but sometimes the target modality can be complex with high heterogeneity.
In future work, we will replace the modality vector by other more flexible modality representations, and develop a style transfer module that can focus on the local difference among different domains.

\subsubsection{Acknowledgement.}
This work was supported by the National Natural Science Foundation of China (61971142), and L. Li was partially supported by the CSC Scholarship. 
JA Schnabel and VA Zimmer would like to acknowledge funding from a Wellcome Trust IEH Award (WT 102431), an EPSRC program Grant (EP/P001009/1), and the Wellcome/EPSRC Center for Medical Engineering (WT 203148/Z/16/Z).

\bibliographystyle{splncs04}
\bibliography{AllBibliography_STACOM2020}

\end{document}